\documentclass[a4paper, 10 pt, journal, twoside]{IEEEtran} 
\IEEEoverridecommandlockouts  
\usepackage[utf8]{inputenc}

\usepackage{graphicx} 
\usepackage{adjustbox}

\usepackage{booktabs} 
\usepackage{array} 
\usepackage{paralist} 
\usepackage{verbatim} 
\usepackage{subfig} 
\usepackage{amsmath}
\usepackage{amsthm}
\usepackage{amssymb}
\usepackage{authblk}
\usepackage[ruled,vlined]{algorithm2e}
\usepackage{cite}
\usepackage{todonotes}
\usepackage{standalone}
\usepackage{bm}
\usepackage{xargs}
\usepackage{tikz}
\usetikzlibrary{decorations.pathreplacing}
\usetikzlibrary{automata,positioning}
\usepackage{tikz-network}
\usepackage{tikz-cd}
\newcommandx{\iman}[2][1=]{\todo[linecolor=orange,backgroundcolor=orange!25,bordercolor=orange,author=Iman,#1]{#2}}

\theoremstyle{plain}
\newtheorem{thm}{Theorem}[section] 

\theoremstyle{definition}
\newtheorem{defn}[thm]{Definition} 
\newtheorem{exmp}[thm]{Example} 

\newtheorem{princ}{Principle}

\usepackage{fancyhdr} 
\newcommand{\Act}{\mathcal{A}}

\newcommand{\Inp}{\Sigma}

\tikzset{commutative diagrams/.cd,
mysymbol/.style={start anchor=center,end anchor=center,draw=none}
}

\DeclareMathOperator*{\seq}{\rightarrow}
\DeclareMathOperator*{\fb}{?}

\DeclareMathOperator*{\prlseq}{\rightrightarrows}

\title{A principled analysis of Behavior Trees and their generalisations}
\author[1]{Oliver Biggar}
\author[1]{Mohammad Zamani}
\author[2]{Iman Shames}
\affil[1]{Defence Science and Technology Group, Australia}
\affil[2]{The Australian National University, Australia}
\date{}

\begin{document}

\maketitle

\begin{abstract}
As complex autonomous robotic systems become more widespread, the need for transparent and reusable Artificial Intelligence (AI) designs becomes more apparent. In this paper we analyse how the principles behind Behavior Trees (BTs), an increasingly popular tree-structured control architecture, are applicable to these goals. Using structured programming as a guide, we analyse the BT principles of \emph{reactiveness} and \emph{modularity} in a formal framework of action selection. Proceeding from these principles, we review a number of challenging use cases of BTs in the literature, and show that reasoning via these principles leads to compatible solutions. Extending these arguments, we introduce a new class of control architectures we call \emph{generalised BTs} or \emph{$k$-BTs} and show how they can extend the applicability of BTs to some of the aforementioned challenging BT use cases while preserving the BT principles.
\end{abstract}
\section{Introduction}

A change from unstructured to structured artificial intelligence (AI) is underway in autonomous cyber-physical systems. That change centres around constructing decision-making in ways which allow AI to be \emph{transparent} and \emph{reusable}. Transparency allows human designers to understand and predict the behavior of autonomous systems, and reusability allows designers to iteratively build behavior of significant complexity.




While the importance of these ideas has been recognised for a long time in AI~\cite{brooks1986robust,nilsson1993teleo}, this change has been slow in cyber-physical systems. The historical bottlenecks of the field have fallen more on the side of the `physical' than the `cyber'---hardware, vision, sensing and the real-time dynamic systems with which they must work closely. By contrast, the world of software (`purely cyber' systems) long ago reached levels of complexity at which the \emph{structure} of programs became the limiting factor in terms of readability, reuse and ease of debugging. The recognition of this problem, and the tectonic shift in the development of software which followed it (now known as the \emph{structured programming movement}), began with Dijkstra's famous 1968 letter `\textbf{go to} statement considered harmful'~\cite{gotoharmful}. There, Dijkstra points out that ``our powers to visualize processes evolving in time are relatively poorly developed. For that reason we should do (as wise programmers aware of our limitations) our utmost to shorten the conceptual gap between the static program and the dynamic process". He argues the tangled structure of programs written with heavy reliance on \textbf{go to} are challenging to understand, and it is difficult to identify and reuse cohesive subparts in new programs. The consequences of this idea influence every modern programming language and allow today's programs to be many orders of magnitude larger in both size and complexity.

Finite State Machines (FSMs) are one of the most common ways of structuring AI for cyber-physical systems. FSMs are `unstructured' in the sense that their graph structure does not closely resemble the processes which they induce. To borrow a quote of Champandard from~\cite{btbook}, ``anyone who has worked with this kind of technology in industry knows how fragile such logic gets as it grows." This analogy between FSMs and \textbf{go to}s has been a common fixture of recent literature on Behavior Trees, a tree-structured control architecture which has recently presented itself as a `well-structured' alternative to FSMs. Unlike FSMs, BTs are \emph{modular} in that every subtree is itself a BT with cohesive behavior, and \emph{reactive}, in that they respond immediately to changes in the environment. The modularity allows subtrees to be easily reused in other trees and the reactiveness make their behavior transparent and predictable. This \emph{reactiveness} and \emph{modularity} has been argued~\cite{btbook,generalise,surveyofbts,UAVmissionBT} as the source of their transparency and ease-of-use~\cite{integratedARM,hu2015semi}. In fact, many authors~\cite{ogren2012increasing,generalise,btbook} have pointed to Dijkstra's letter as an analogy for the use of BTs over FSMs.

 BTs---aiming to fill this `structured AI' niche---first appeared in the control of non-player characters in computer game AI. In anticipation of its growing importance for physical autonomous systems, the last ten years has seen significant transitions of BT research from the game AI universe into that of robotics (for a recent survey, see~\cite{surveyofbts}).

With the community of `Behavior Trees in robotics' research being now roughly a decade old, it seems a reasonable point to pause and reflect upon what seems to have worked and what has not. While BTs (and similar `structured architectures') have been praised for their modularity and reactiveness, there have also been cases where their use seems unwieldy~\cite{anguelovBTmisuse,modularity}. Why and when this occurs is what we wish to answer within this paper.

Nowadays, the principles of structured programming are ubiquitous (and form the basis for more complex programming paradigms, such as object-oriented programming), and yet few programming languages are `purely' structured, in the sense of~\cite{dijkstra1970notes}. Most instead contain some exceptions to the `rules', usually in the form of early exit from loops (by \textbf{break} statements) and from functions (by \textbf{return} statements). In 1974, Knuth~\cite{knuth1974structured} predicted this development, stating that while reducing the use of \textbf{go to} statements was in many cases desirable, certain constructions still required it---`structuring' them led to more obscure and opaque code. The benefits of structured programming still remain---the challenge is to identify precisely when those benefits can be exploited. If structured approaches fail, one must determine the next-best solution.

In this paper, we aim to study Behavior Trees in robotic AI in this way. That is, we isolate the fundamental principles of BTs, \emph{reactiveness} and \emph{modularity}, and we look at how these principles allow for reusable and transparent behavior. Then, we provide a number of examples from the robotics literature and analyse where and how BTs should be applied so as to maximise their benefits.  

We argue that, much as for \textbf{go to} statements, the real need for `unstructured AI' is rare, but it does occur and we should recognise it. We should not becessarily attempt to simply rewrite our FSMs as BTs, when doing so results in AI which is no clearer~\cite{expressiveness}. As Dijkstra~\cite{gotoharmful} states about the structured program theorem: ``The exercise to translate an arbitrary flow diagram more or less mechanically into a jump-less one, however, is not to be recommended. Then the resulting flow diagram cannot be expected to be more transparent than the original one." Or, consider Knuth's words~\cite{knuth1974structured}: ``Probably the worst mistake any one can make with respect to the subject of \textbf{go to} statements is to assume that `structured programming' is achieved by writing programs as we always have and then eliminating the \textbf{go to}'s." Instead, we attempt to learn, from architectures like Behavior Trees, how we should structure AI in ways that lead to transparency and reusability. We believe the lack of attention to these is a potential cause of the ``widespread misuse" of BTs outlined by Anguelov~\cite{anguelovBTmisuse}, which he argues has led to ``monolithic trees whose size and complexity has made it all but impossible to extend or refactor without the risk of functional regression."

We will discuss a series of questions surrounding BT use: memory and mechanisms for storing knowledge in BTs; modeling Success and Failure of Actions; extensions of BT concepts; expressing strategy in BTs;  encapsulating Action metadata for reuse in BTs. Our goal is for this discussion to serve as an aid in determining when a BT could be an important part of a robotic architecture. Finally, we show how a natural series of assumptions allows us to make the most of BTs' reactiveness and modularity while increasing the breadth of behavior they can express. The resultant architecture, which we call a \emph{generalised BT} or $k$-BT, allows for any finite number ($k$) of return values and associated control flow nodes. $k$-BTs are naturally suited to situations where the `Success' and `Failure' nodes do not conceptually fit or are too limiting.

 The rest of the paper is organised as follows. Section~\ref{sec:related work} discusses related work. . Section~\ref{sec:asms} introduces our definitions and assumptions. Section~\ref{sec:classicalbts} introduces BTs and FSMs formally, and briefly mentions Decision trees (DTs) and Teleo-Reactive programs (TRs)~\cite{nilsson1993teleo}. In Section~\ref{sec:btprinciples} we discuss the two main principles of BT use: reactiveness and modularity. Section~\ref{sec:practical} then discusses a number of use cases for BTs and provide solutions motivated by the aforementioned principles. Section~\ref{sec:kbts} introduces $k$-BTs and their application towards some of the problems previously discussed. Finally, Section~\ref{sec:conclusions} concludes the paper.

 \section{Related Work} \label{sec:related work}
 
 This paper was written concurrently with~\cite{modularity} and~\cite{expressiveness}, and shares many structural features with both frameworks, particularly the latter. This paper contains the material that is focused largely on BTs and their use, while~\cite{modularity} focuses on modularity in a number of reactive architectures and~\cite{expressiveness} focuses on comparing the expressiveness of different control architectures, including BTs. While we discuss $k$-BTs in~\cite{modularity}, this is the paper in which we formally introduce them.
 
 Many of the goals of our work have been explored to varying degrees in previous work. Both modularity and reactiveness have previously been described as principles of BT use, and these are discussed at in~\cite{btbook,generalise,surveyofbts,UAVmissionBT}. We build on this discussion, and particularly show how this applies to practical BT use cases.
 
 The discussion of BT use in robotics is most similar in style and content to~\cite{btbook}, and we take a number of examples from this text. However, we also differ in some of the conclusions we form from the principles. 
 Anguelov previously set out some of the challenges of BT use and misuse in~\cite{anguelovBTmisuse}. His focus is on BTs in game AI, where issues such as implementation and efficiency are more important, but some of the high-level ideas are similar.
 

\section{Assumptions and system structure}\label{sec:asms}

 In this paper we consider BTs, FSMs, DTs and TRs as devices for switching between actions or modes of operation in a robotic system. We present a formal discussion of this in~\cite{expressiveness}, but in this paper we focus only on general ideas and omit the formalism. We will generally refer to the world with which the robot interacts as its \emph{environment}. These architectures determine their action-selection in response to \emph{input}, which is some abstract information. The nature of what we consider to be input is important, and we will discuss the consequences of such assumptions on a case-by-case basis later---particularly see Section~\ref{sec:reactiveness}. For now though, we point out only that we do not require input to be `input from the robot's sensors' specifically, as in general input may be highly abstracted. This is because we do not expect BTs (or indeed FSMs, DTs or TRs) to comprise the entirety of the robotic system under consideration. By allowing a flexible interpretation of input we can focus our attention on the parts of the decision-making that are the responsibility of the BT. We will make this easier by assuming the decision making of our robot is structured in \emph{layers}.

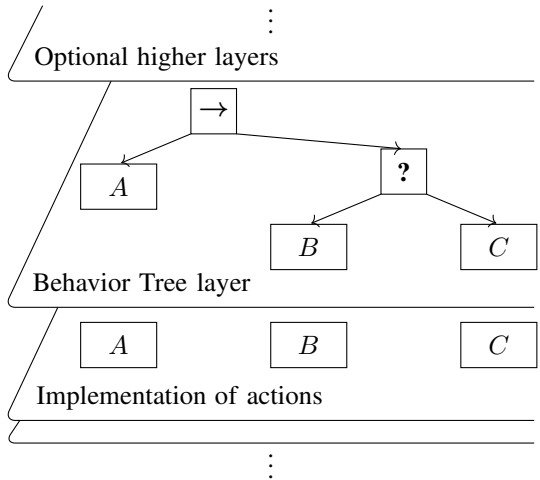
\begin{figure}
    \centering
    \begin{tikzpicture}[shorten >=1pt,node distance=2cm,on grid,auto] 
    \draw[rounded corners=4pt](1.4,4.5) -- (0,1.5) -- (7,1.5);
    \draw[rounded corners=4pt] (0.5,5.5) -- (0,4.5) -- (7,4.5);
    \draw[rounded corners=4pt] (0.7,1.5) -- (0,0) -- (7,0);
    \draw[rounded corners=4pt] (0.2,0) -- (0,-0.3) -- (7,-0.3);
    \node at (3.5,-0.5) {$\vdots$};
    \node at (3.5,5.4) {$\vdots$};
    \node at (2.3,0.3) {Implementation of actions};
    \node at (1.8,1.8) {Behavior Tree layer};
    \node at (2,4.8) {Optional higher layers};
    \node at (1.5,1) {$A$};
    \node at (4,1) {$B$};
    \node at (6.5,1) {$C$};
    \draw (1,0.7) rectangle ++(1,0.6);
    \draw (3.5,0.7) rectangle ++(1,0.6);
    \draw (6,0.7) rectangle ++(1,0.6);
    
    \draw (1,2.8) rectangle ++(1,0.6);
    \draw (3.5,2) rectangle ++(1,0.6);
    \draw (6,2) rectangle ++(1,0.6);
    
    \draw (4.95,3) rectangle ++(0.6,0.6);
    \node at (5.25,3.3) {\textbf{?}};
    \draw (2.45,3.8) rectangle ++(0.6,0.6);
    \node at (2.75,4.1) {$\bm{\seq}$};
    \draw[->] (2.45,3.8) -- (1.5,3.4);
    \draw[->] (3.05,3.8) -- (5.25,3.6);
    \draw[->] (4.95,3) -- (4,2.6);
    \draw[->] (5.55,3) -- (6.5,2.6);
    \node at (1.5,3.1) {$A$};
    \node at (4,2.3) {$B$};
    \node at (6.5,2.3) {$C$};
    \end{tikzpicture}
    \caption{A layered robotic control architecture}
    \label{fig:layers}
\end{figure}

Layering is a standard concept in robotic and software design, as it separates concerns at different levels of abstraction. One assumes that higher layers act upon the layers beneath, but not the other way around. The part of our robotic architecture which is controlled by a BT we shall therefore call the \emph{BT layer} (see Fig~\ref{fig:layers}). The layers below the BT correspond to controllers for the individual actions from which the BT selects. These could be continuous controllers operating on actuators or possibly themselves complex action selection mechanisms operating on yet lower layers---these too could be implemented as BTs \footnote{The BT layer need not be truly a single layer. If BT implementations of layers appear distinctly in a robotic architecture then the same principles will apply to both and we will refer to either as `the' BT layer. Note that if a BT selects a BT in an immediately lower layer these BTs can be joined to a single larger BT, by the composition properties of BTs.}. Likewise, there may be layers above the BT, which select this BT under various circumstances, and these again could also be BTs. BTs are also described in this way in~\cite{unifiedframework}. By making this `layered architecture' assumption, we can more clearly specify the focus of our investigation in this paper, which is to understand which aspects of the decision-making are best structured within the BT layer, and how this can be done as clearly as possible.

\section{Finite State Machines, Decision Trees, Teleo-reactive programs and Behavior Trees} \label{sec:classicalbts}
These definitions are in general the same as those in~\cite{modularity} and~\cite{expressiveness}.

\subsection{Teleo-reactive programs}

Teleo-reactive programs (TRs)~\cite{nilsson1993teleo} are lists of condition-action rules.\begin{align*}
    k_1 &\to a_1 \\ 
    k_2 &\to a_2 \\
    &\vdots \\
    k_n &\to a_n 
\end{align*} They are executed by continuously scanning the list of conditions $k_i$ in the order of their priority, and executing the action $a_i$ associated with the first satisfied condition. As these conditions become true and false, the selected action changes immediately to the action corresponding to the first satisfied condition. The \emph{teleo} indicates that such lists are goal-oriented while \emph{reactive} is intended to describe how they react constantly to changes in the environment. Reactiveness in this sense is a central principle of this paper.

We note that Nilsson's original TRs were equipped with somewhat more functionality, being able, for instance, to have their conditions and actions take arguments. Here, as in~\cite{btbook} we treat them in their simplest sense as condition-action lists.
\subsection{Decision Trees}
Decision Trees (DTs)~\cite{btbook} are decision-making tools which resemble the structure of if-then statements. Specifically, DTs are binary trees where leaves represent actions and internal nodes represent predicates. The two arcs out of each predicate are labelled by `True' and `False'. Execution of DTs occurs by beginning at the root and evaluating each predicate on the current input state until a leaf is reached, at which point that action is executed. At a predicate node, if it is true in the current input state the execution proceeds down the `True' arc, and otherwise down the `False' arc. Like TRs, DTs are executed by continuously checking the predicates against the current state of the world.
\subsection{Behavior Trees}
BTs are control architectures which take the form of ordered directed trees. The execution of a BT occurs through signals called `ticks', which are generated by the root node and sent to its children. A node is executed when it receives ticks. Internal nodes tick their children when ticked, and are called \emph{control flow nodes} and leaf nodes are called \emph{execution nodes}. When ticked, each node can return one of three possible return values; `Success' if it has achieved its goal, `Failure' if it cannot operate properly and `Running' otherwise, indicating its execution is underway. Typically, there are four types of control flow nodes (Sequence, Fallback, Parallel, and Decorator) and two types of execution node (Action and Condition)~\cite{btbook}. Note that Fallback is sometimes called Selector. A Condition (drawn as an ellipse) checks some proposition (based on the current input), returning Success if true and Failure otherwise. An Action node (drawn as a rectangle) represents an action taken by the agent. Sequence nodes (drawn as a $\seq$ symbol) tick their children from left to right. If any children return Failure or Running that value is immediately returned by the Sequence node, and it returns Success only if every child returns Success. Fallback (drawn as a $\fb$) is analogous to Sequence, except that it returns Failure only if every child returns Failure, and so on. The Parallel node (symbol $\prlseq$) has a success threshold $M$, and ticks all of its $N$ children simultaneously, returning Success if $M$ of its children return Success, Failure if $N-M +1$ return Failure and Running otherwise. The Decorator node returns a value based on some user-defined policy regarding the return values of its children. For a more detailed discussion, we refer the reader to~\cite{btbook}. Following~\cite{biggar2020framework}, we may write BTs succinctly in infix notation, such as $A\seq (B \fb C) \seq D$ where $A,B,C,D$ are interpreted as leaf nodes and the control flow nodes $\seq$ and $\fb$ are interpreted as associative operators over the leaf nodes.
\subsection{Finite State Machines}
A Finite State Machine (FSM) is a control architecture structured as a labelled directed graph, where nodes represent \emph{states}. The machine is in exactly one of these states at a given time. Arcs, called \emph{transitions}, link states from one to another. A transition from one state to another is undertaken in response to input which \emph{triggers} the transition. There is a single state called the initial state. States are labelled by actions. FSMs execute by beginning in the initial state, and whenever input is received, taking any transitions which are triggered by that input, until reaching a state where no transitions are triggered. The FSM then executes the action labelling that state until new input is received (i.e. there is an observable change in the environment)  Formally, an FSM is a six-tuple $(Q,q_0,\Inp,\Act,\delta,\ell)$, where $Q$ is a finite set of states, $q_0\in Q$ is the initial state, $\Inp$ is a finite set called the input alphabet, $\Act$ is a finite set of actions, $\delta:\Inp\times Q\to Q$ is the transition function dictating when transitions are triggered and $\ell:Q\to \Act$ is the output function assigning actions to states. This particular formal interpretation of a Finite State Machine is called a \emph{Moore machine}\cite{moore1956gedanken}.

\section{The principles of Behavior Tree use} \label{sec:btprinciples}

In order to reason about when (at which layer of decision making) BTs should be used, we must know why they should be used at all. In general when explaining why BTs are useful, the arguments which are recurrent in BT literature are the following:
\begin{itemize}
    \item Behavior Trees are reactive~\cite{btbook,UAVmissionBT,sprague2018adding,biggar2020framework,modularity}
    \item Behavior Trees are modular~\cite{ogren2012increasing,btbook,generalise,UAVmissionBT,sprague2018adding,biggar2020framework,modularity}
\end{itemize}
Fundamentally, these properties make BTs easy to construct, analyse, read, reuse and debug. Some authors have additionally described  \emph{reusability}~\cite{isla2015handling,generalise,ogren2012increasing,biggar2020framework} or \emph{readability}~\cite{isla2015handling,integratedARM,UAVmissionBT} as principles of BT use. While we agree with their fundamental importance, we shall view these as the \emph{end} to which reactiveness and modularity are a \emph{means}. That is, readability and reusability are considered as the goals of `structured robotic AI', and modularity and reactiveness the tools by which BTs specifically achieve those goals. Now we unpack both in more depth.

\subsection{Reactiveness} \label{sec:reactiveness}
Intuitively, reactiveness means that when the environment changes the system should always respond appropriately---that is, it should be \emph{reactive} to changes in the input. In the BT case, this is achieved through the repeated ticking, which frequently polls the environment for its current state. In describing his Teleo-\emph{reactive} programs, Nilsson describes this property as ``circuit semantics"~\cite{nilsson1993teleo}. This inspires our definition, which is the following:

\begin{defn} \label{def:reactive}
An architecture is \emph{reactive} if its decision-making depends only on the \emph{current} state of the environment.
\end{defn}

In~\cite{btbook}, the authors describe \emph{reactiveness} as desirable because long chains of open-loop actions are often problematic in unstructured and unpredictable environments. BTs, by contrast, check the current state of the environment every tick and respond most appropriately to all changes. As an example, consider a robot which must locate and pick up a block, then carry and deposit that block in a bin. If while transporting this block the robot loses its grip, a \emph{reactive} architecture would again attempt to locate and grasp the block, while a poorly-designed unreactive architecture would continue towards the bin\footnote{It is not \emph{required} that a non-reactive architecture behave in this way---clearly reactive architectures are a subset of all architectures, so these `reactive' behaviors can be constructed in non-reactive architectures. However, the purpose of a reactive architecture is to make avoiding these mistakes simple. In a FSM, transitions must be explicitly constructed for many possibilities  that are handled implicitly by reactive architectures.}.

There is an important subtlety here, discussed extensively in~\cite{expressiveness}, which is that reactiveness depends on what constitutes input. In the BT literature~\cite{btbook}, FSMs are viewed as not being reactive in general, with the above example as an argument for why. This is because while FSMs always respond to pairs $(x,q)$ of input and state with an action, the current state $q$ is usually not counted as `input'. The reason for this is that $q$ is not considered outwardly `visible' to the user.

Reactiveness makes reasoning about BTs extremely intuitive; given some external world state as input, it is always straightforward to determine which action will be selected in response. Reactive control architectures can be thought of as a partition of the input space, where regions of the partitions are labelled by the action selected in that region. Many authors have argued that reactiveness is at the core of BT use~\cite{btbook,sprague2018adding,biggar2020framework,modularity}, and that a primary goal of BTs is to add a reactive layer to a decision making architecture. From this, we state the following principle.

\begin{princ}
Behavior Trees should be reactive; whenever the current input is the same, they should always select the same action regardless of input history.
\end{princ}

Reactiveness can be thought of as \emph{memorylessness}. When considered in this way, we can think of reactive behaviors like instinctive or unconscious behaviors in a biological system. The limited computational load required to process BTs forms a core part of their readability, and their applicability to a variety of systems, including those with very limited computing power. Of course, we achieve this simplicity and clarity by restricting the access of an architecture to memory, so this does mean BTs are less expressive than more complex architectures like FSMs~\cite{expressiveness}.



We propose the following test for the property of reactiveness. Consider a robot in an empty room, and a human observer. An architecture is reactive if, when provided with the same stimulus, the robot always reacts in the same way. This is fundamentally tied to its behavior being \emph{predictable} and \emph{transparent}. The human observer should always be able to predict the robot's response to any given stimulus, given a representation of its architecture, without any knowledge of what prior stimulus the robot had been exposed to. This test will be useful for highlighting cases where reactiveness is violated, such as the FSM case discussed above.

\subsection{Modularity} \label{sec:modularity}

The modularity of BTs means that they can be composed, decomposed and reused because subtrees and actions are themselves well-defined BTs. This modularity comes about both from structure (being a recursively defined tree structure) and having a fixed interface given by three return values. Modularity is widely regarded as key to the use of BTs~\cite{ogren2012increasing,btbook,generalise,UAVmissionBT,sprague2018adding,biggar2020framework,modularity}, as it allows libraries of BTs to be created which can be easily combined. Any constructed subtree of a BT can be reused in any other tree. From this, we propose the following principle.

\begin{princ}
Behavior Trees should be modular; every subtree, including the whole tree, should be able to be reused meaningfully in another tree.
\end{princ}

\section{Applying these principles in practice}
\label{sec:practical}
We examine now how these principles can help us to judge cases of BT use in practice. In this section we show that interpreting these principles provides insights or solutions to a number of BT challenges. The examples used in this section are inspired by examples used in the BT literature. Structurally, we split this section by problems which primarily address reactiveness or modularity. This partition is not strict; many of the points made are relevant to both principles.
\subsection{Reactiveness}
\subsubsection{Implicit memory} \label{sec:implicitmemory}

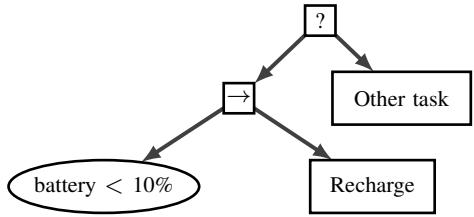
\begin{figure}
    \centering
    \begin{tikzpicture}
\clip (0,0) rectangle (7.0,3.0);
\Vertex[x=4.388,y=2.600,size=0.4,opacity=0,label=?,fontscale=1.2,shape=rectangle,style={}]{1}
\Vertex[x=3.318,y=1.571,size=0.4,opacity=0,label=$\rightarrow$,fontscale=1.2,shape=rectangle,style={}]{2}
\Vertex[x=1.542,y=0.400,size=0.7,opacity=0,label=battery $<$ 10\%,fontscale=1.2,shape=ellipse,style={minimum width = width("batterysls10p")+4ex}]{3}
\Vertex[x=5.074,y=0.400,size=0.7,opacity=0,label=Recharge,fontscale=1.2,shape=rectangle,style={minimum width = width("Recharge")+2ex}]{4}
\Vertex[x=5.458,y=1.571,size=0.7,opacity=0,label=Other task,fontscale=1.2,shape=rectangle,style={minimum width = width("Otherstask")+2ex}]{5}
\Edge[,fontscale=1.2,Direct](1)(2)
\Edge[,fontscale=1.2,Direct](1)(5)
\Edge[,fontscale=1.2,Direct](2)(3)
\Edge[,fontscale=1.2,Direct](2)(4)
\end{tikzpicture}
    \caption{A BT which recharges a robot when the battery level falls below 10\%.}
    \label{fig:impl1}
\end{figure}

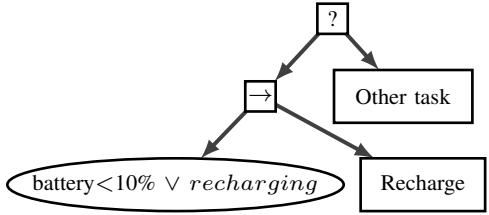
\begin{figure}
    \centering
    \begin{tikzpicture}
\clip (0,0) rectangle (7.0,3.0);
\Vertex[x=4.418,y=2.600,size=0.4,opacity=0,label=?,fontscale=1.2,shape=rectangle,style={}]{1}
\Vertex[x=3.470,y=1.571,size=0.4,opacity=0,label=$\rightarrow$,fontscale=1.2,shape=rectangle,style={}]{2}
\Vertex[x=2.353,y=0.400,size=0.7,opacity=0,label=battery$<$10\% $\lor$ $recharging$,fontscale=1.2,shape=ellipse,style={minimum width = width("battery$<$10\% $\lor$ $recharging$")+2ex}]{3}
\Vertex[x=5.606,y=0.400,size=0.7,opacity=0,label=Recharge,fontscale=1.2,shape=rectangle,style={minimum width = width("Recharge")+2ex}]{4}
\Vertex[x=5.347,y=1.571,size=0.7,opacity=0,label=Other task,fontscale=1.2,shape=rectangle,style={minimum width = width("Otherstask")+2ex}]{5}
\Edge[,fontscale=1.2,Direct](1)(2)
\Edge[,fontscale=1.2,Direct](1)(5)
\Edge[,fontscale=1.2,Direct](2)(3)
\Edge[,fontscale=1.2,Direct](2)(4)
\end{tikzpicture}
    \caption{A second BT for recharging, which avoids chattering with an auxiliary variable}
    \label{fig:impl2}
\end{figure}

Consider the BT of Figure~\ref{fig:impl1}, assumed to be controlling a robot. 
When the robot's battery is less than 10\% it recharges, and otherwise it performs some task. The issue is, implemented as shown, we can have chattering where the robot fluctuates rapidly between Recharge and Other task as the battery level fluctuates around 10\%. It is tempting to change this to the BT of Figure~\ref{fig:impl2}.
Here we have introduced an auxiliary variable $recharging$ which becomes true after we begin recharging and becomes false once the battery reaches 100\%. Now the robot remains doing `Recharge' until its battery is full. However, in doing this, we have violated the principle of reactiveness, and lost transparency. Consider some input where the battery level is $>10\%$. An observer cannot predict how the robot will behave in this state without whether it had been Recharging \emph{in the past}. This auxiliary variable has introduced what we shall call \emph{implicit memory}. As for the Finite State Machine example given in Section~\ref{sec:reactiveness}, the variable $recharging$ cannot be considered part of the input because it is not externally visible. We cannot reason about this variable as we would about a property of the environment, and so the operation of the tree is no longer transparent. This variable cannot take different values at any point---it becomes true and false under prescribed conditions, but the information of these conditions is not contained in the above BT; we had to explain in the text that $recharging$ becomes False when the battery level reaches 100\%, because this could not be derived from the diagram. This breaks the cohesiveness of the architecture, and from a practical perspective means that manipulation of this variable must also be implemented outside the normal software framework for the BT. This can easily lead to errors.

\begin{figure}
    \centering
    \begin{tikzpicture}[shorten >=1pt,node distance=2cm,on grid,auto] 
    \draw[rounded corners=4pt](0.7,1.5) -- (0,0) -- (8,0);
    \draw[rounded corners=4pt] (0.5,2.5) -- (0,1.5) -- (8,1.5);
    \node at (1.8,0.3) {Behavior Tree layer};
    \node at (2,3.3) {};
    \node[initial,state] (D) at (2,3) {Other Task};
    \node[state] (R) at (6.5,3) {Recharging};
    \path[->]
        (D) edge[out=40] node {$battery <10\%$} (R) ;
    \path[->]
        (R) edge[out=200,in =-20] node {$battery=100\%$} (D) ;
    \draw (1.5,0.6) rectangle ++(3,0.7);
    \node at (3,0.9) {Other Task BT};
    \end{tikzpicture}
    \caption{A solution to the chattering problem.}
    \label{fig:chattering}
\end{figure}
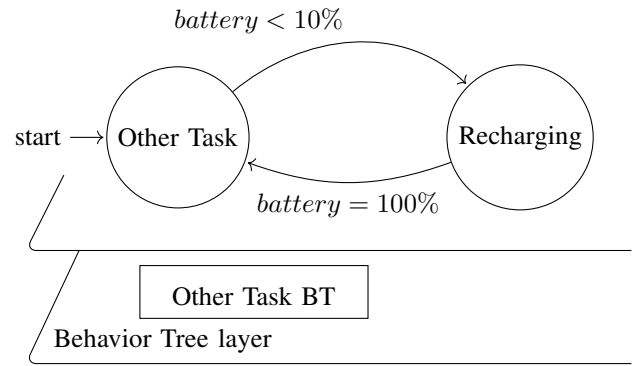


How instead might we handle challenges such as this chattering problem? Here, the issue stems from the fact that the robot should \emph{not} be reactive while it is charging. To be truly reactive, architectures must be able to interrupt their children in response to any new stimulus from the environment. In other words, the BT layer can always interrupt all layers beneath it. If we instead require actions which will never be interrupted, we propose placing them at a layer above the BT layer (or any reactive layer). We show this solution in Figure~\ref{fig:chattering}, where an FSM controls switching in and out of a recharging state. When not recharging, the BT controlling `Other task' is operated on the layer beneath. The overall architecture is still not reactive, but this approach clearly separates the reactive and non-reactive layers and the transition conditions for $recharging$ are now transparent, improving readability.

We used this example to introduce the challenges of implicit memory, but using implicit memory is not the only proposed solution to this specific chattering problem in the BT literature. For example,~\cite{UAVmissionBT} uses a notion called \emph{transient tasks} to address this problem. This example was adapted from~\cite{generalise}. A similar problem is discussed in~\cite{expressiveness}.

\subsubsection{Modelling Success and Failure} \label{sec:modellingreturnvalues}

When designing BTs or abstracting some implemented behavior into actions, one must decide on the conditions under which actions should succeed or fail. We shall call these conditions the \emph{return conditions}. We show now that we cannot do this freely if we are to preserve reactiveness---in a sense, these conditions must also be `reactive'.

To begin, consider an action `Send Message'. When should this return Success or Failure? Requiring that it returns Success if `a message has been sent' is problematic, because it introduces implicit memory, violating reactiveness. To see this, note that this condition is stated \emph{in past tense}---it refers to what has happened previously. We cannot evaluate its truth or falsity without memory. Just as with previous example of implicit memory, the variable storing when `a message has been sent' is not defined within the tree, and is controlled at some other layer, breaking transparency. An observer cannot predict the BT selection without knowing of the value of this variable, and it is not externally visible `input'. Consider a more extreme example, where an action returns Success if a given sequence of variables has been true in the past. Using such actions would allow us to construct arbitrarily complex behaviors, so must violate reactiveness by the results of~\cite{expressiveness}. These same ideas apply to conditions. Consider an example of an agent who must pass through a door, with action `Walk Forward' and a condition `Agent Has Passed' checking when the agent reaches the next room. The problem with this condition is in its past-tense description---if the agent is simply in the next room, there is no way of knowing if it did or did not pass through the door in the past.

This problem can occur easily, but can also often be solved by rewriting conditions in present tense, or omitting them if no present-tense conditions are reasonable. Consider an action `Grasp Object'. One intuitive way of modelling its return conditions is to say it returns Success if an object has been grasped and failure if it cannot grasp an object. However, this requires implicit memory to record whether we have already attempted to grasp the object. Luckily, we can avoid the problem completely by having Grasp Object return Success if the robot is holding an object and Failure if there is no object within reach. Thus, if there is an object within reach and not grasped, the robot executes a grasp operation until either it is holding the object or the object is no longer present. Both return conditions are both present-tense, and refer to observable states. Recall the motivating examples for reactiveness discussed earlier, where a robot grasping a ball drops it while completing a task, and immediately reacts by searching for the ball~\cite{btbook}. This requires that `Grasp Ball' does not alwasy return Success when a ball was grasped in the past. One way of thinking about the distinction between these models is that in a reactive architecture we care about outcomes not processes---it matters only that we are now holding an object, not whether we actually used the grasp action to achieve that outcome. Likewise, the `Agent Has Passed' condition could be replaced with `Agent in Room X', a present-tense condition checking the current location. This present-tense check also is simpler.

In arguing that `a message was sent' was implicit memory, we assumed that this variable is not outwardly visible, and cannot be interpreted as sensor input to the BT. Another solution to this problem is to rephrase this in a way that \emph{can} be interpreted as an input. For instance, suppose Send Message returns Success if the robot observes a human operator nodding, or another agent receiving the message provides an acknowledgement of its receipt. Depending on circumstances, this might be considered visible (though this is a grey area---see Section~\ref{sec:conditions requiring action}). This option at least avoids the need for implicit memory.

One common source of these examples is discrete or non-durative actions. A simple case is the action Step Forward for a BT operating in a discrete grid world. How should we assign a Success condition to this action? We should avoid success conditions along the lines of `Success if we have stepped forward' or `Success if we are our position is one forward from our last position', because, again, they are past-tense. We postulate that there is no reactive way to assign a Success conditions to such actions---a better solution is to never return Success. Instead the agent should step forward repeatedly until the BT switches tasks or some present-tense Failure condition becomes true, such as $Path Obstructed$. This example is also explored in Section~\ref{sec:atomic actions}, where we argue that very low-level discrete actions are hard to use in BTs.

The takeaway of this section is that for a BT to be reactive, the Success and Failure conditions of each action must also be reactive. That is, the return value of all actions must be determined only by current state.

The Sent Message example was adapted from~\cite{biggar2020framework}, and the door example was adapted from~\cite{btbook}. A similar example using waypoints is discussed in~\cite{UAVmissionBT}. We give a similar discussion in~\cite{expressiveness}.

\subsubsection{Knowledge versus memory} \label{sec:conditions requiring action}

In the last section we concluded that actions should be able to return Success or Failure on the basis of current input only. In other words, the return value should not be based on whether `something was done' but rather whether `something is true'. There are, however, cases where this thinking itself introduces additional challenges.

Consider an action Unlock Door. Some Success and Failure conditions for this action might be the variables $Door Unlocked$ and $NoKey$ respectively. However, from the robot's point of view, whether a closed door is locked or unlocked may not be clear from input. It is too generous to assume this information is provided when such doors appear identical. In order to determine the value of $DoorUnlocked$, the robot must first try some action, say, attempting to unlock the door. In other words, Unlock Door might return Running even though its Success condition is in fact satisfied, depending on whether or not it is aware of this, which seems to violate reactiveness.

There are a few possible solutions to this problem. One is that we assume that all return conditions are based only on unambiguously visible inputs, thus rejecting any ambiguous cases of this form. However, this does impose significant limits on the operation of BTs. We recommend a more practical compromise: we require the robot to be reactive \emph{in the context of its own world view}. In other words, from the BT perspective we allow $Door Unlocked$ to be either true, false or unknown, so the decisions of the BT are still reactive on this basis. 

We must therefore make an important but subtle distinction between the concept of knowledge, which is the agent's representation of the world, and memory, which is the agent's representation of that world's \emph{history}. The former is essential for robotic applications, but reactiveness argues that the latter is not (always). This internal world model abstracts sensor input into a cohesive structure which can be queried by Behavior Trees. Doing so is consistent with reactiveness, so long as only the current state of this model is used to determine the action selected. This world view need not be static, and can update itself as new information is received, such as in the door-unlocking example used above. We do instead require that this internal model be merely a representation of the external world, and not a blackboard for \emph{any} information to be stored. If that were the case, variables such as $Recharging$ could be stored there, allowing us the freedom to use implicit memory. Balancing these requirements can be difficult. We judge the overall approach using our test for reactiveness, but where we assume that the observer's world view is the same as that currently possessed by the robot. In this context, we require that robot's behavior still be transparent given that the observer knows the robot does not have full information. A solution similar to this is suggested in~\cite{isla2015handling}.

Finally, we also note that while we should not use an internal world model as a store for arbitrary memory, we do allow, and indeed expect, that agents use the physical world as a store of memory. That is, if a robot opens a door and moves through it, then later returns to that same doorway and finds the door still open, it will move through it without reopening. This may seem obvious, but it is important. If the robot records every doorway it passes through by marking them with chalk, it could, in theory, retrace its steps by following doorways marked with chalk. Note that the robot has not remembered its path, but rather is reactively following the doors marked with chalk. While we argue against an `internal blackboard' because it is not transparent, this `external blackboard'\footnote{Rather literally.} is transparent from the perspective of an outside observer and only depends on the present-tense conditions. If another agent added or removed these marks the robot would respond appropriately.

This example was adapted from~\cite{btbook}.

\subsubsection{Nodes with memory} \label{sec:nodeswithmemory}

We have seen above a number of BT challenges where the problem is caused the lack of access of a BT to memory. In fact, we defined `reactive architectures' as architectures satisfying a particular condition, so it may not be surprising that there is a trade-off between expressiveness (in terms of constructing numerous behaviors) and reactiveness~\cite{expressiveness}. Indeed, though we argue that reactiveness is useful for constructing simple and trustable AI, we are unlikely to construct an architecture where every layer is reactive. Given that memory is generally necessary, how should we incorporate it, keeping in mind our principles and goals?

Firstly, given that we aim to construct transparent architectures, memory use should be explicit and clear. From the perspective of the BT layer, the simplest way to achieve this is to encapsulate memory within individual actions, or in a higher layer controlling access to the BT. Memory use can be readable if it used judiciously.

Unfortunately even this clean separation can be difficult to achieve, as sometimes the BT layer must work closely with layers using memory. One way to do this transparently, suggested in the BT literature, is \emph{control flow nodes with memory}.

Nodes with memory~\cite{btbook} are an extension of the usual BT framework designed to prevent unnecessary ticking in circumstances where reactiveness is not desired. Two variants on the usual control flow nodes are introduced, called \emph{Sequence and Fallback with memory}, and written $\seq^*$ and $\fb^*$. The Sequence node with memory ticks its children from left to right, but only until they return Success, after which it remembers that value without ticking the child again. The memory is cleared once the node with memory itself returns a value other than Running, and on the subsequent tick it begins again with the leftmost node. This is not reactive, but it does at least prevent uncontrolled access to memory. Its operation is quite readable and is still modular, as it retains the tree structure and interface of BTs. The Fallback node with memory operates similarly. 

By the modularity of BTs, we can think of the node with memory and its subtree as being a non-reactive layer beneath the BT layer. The node with memory enforces a specific structure on that lower layer, which allows them to be presented as one layer in a readable fashion. Used judiciously, such nodes are fairly innocuous (however we show in~\cite{expressiveness} that BTs equipped with such nodes are still strictly less expressive than Finite State Machines).

Note that a Sequence with memory
\begin{figure}
    \centering
    \begin{tikzpicture}
\clip (0,0) rectangle (4.0,2.0);
\Vertex[x=1.988,y=1.600,size=0.6,opacity=0,label=$\rightarrow^*$,fontscale=1.2,shape=rectangle,style={}]{1}
\Vertex[x=0.929,y=0.400,size=0.7,opacity=0,label=Action1,fontscale=1.2,shape=rectangle,style={minimum width = width("Action1")+2ex}]{2}
\Vertex[x=3.071,y=0.400,size=0.7,opacity=0,label=Action2,fontscale=1.2,shape=rectangle,style={minimum width = width("Action2")+2ex}]{3}
\Edge[,fontscale=1.2,Direct](1)(2)
\Edge[,fontscale=1.2,Direct](1)(3)
\end{tikzpicture}
    \caption{A Sequence node with memory.}
    \label{fig:seqmem1}
\end{figure}
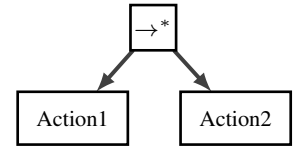
can be emulated by a normal Sequence by adding extra variables~\cite{btbook}.
\begin{figure}
    \centering
    \begin{tikzpicture}
\clip (0,0) rectangle (8.0,3.0);
\Vertex[x=4.05,y=2.600,size=0.4,opacity=0,label=$\rightarrow$,fontscale=1.2,shape=rectangle,style={}]{1}
\Vertex[x=3.069,y=1.571,size=0.4,opacity=0,label=?,fontscale=1.2,shape=rectangle,style={}]{2}
\Vertex[x=1.106,y=0.400,size=0.7,opacity=0,label=action1Done,fontscale=1.2,shape=ellipse,style={minimum width = width("action1Done")+2ex}]{3}
\Vertex[x=3.069,y=0.400,size=0.7,opacity=0,label=Action1,fontscale=1.2,shape=rectangle,style={minimum width = width("Action1")+2ex}]{4}
\Vertex[x=5.031,y=1.571,size=0.4,opacity=0,label=?,fontscale=1.2,shape=rectangle,style={}]{5}
\Vertex[x=5.031,y=0.400,size=0.7,opacity=0,label=action2Done,fontscale=1.2,shape=ellipse,style={minimum width = width("action2Done")+2ex}]{6}
\Vertex[x=6.994,y=0.400,size=0.7,opacity=0,label=Action2,fontscale=1.2,shape=rectangle,style={minimum width = width("Action2")+2ex}]{7}
\Edge[,fontscale=1.2,Direct](1)(2)
\Edge[,fontscale=1.2,Direct](1)(5)
\Edge[,fontscale=1.2,Direct](2)(3)
\Edge[,fontscale=1.2,Direct](2)(4)
\Edge[,fontscale=1.2,Direct](5)(6)
\Edge[,fontscale=1.2,Direct](5)(7)
\end{tikzpicture}
    \caption{A BT which is equivalent to that in Figure~\ref{fig:seqmem1}, using implicit memory rather than a memory node.}
    \label{fig:seqmem2}
\end{figure}
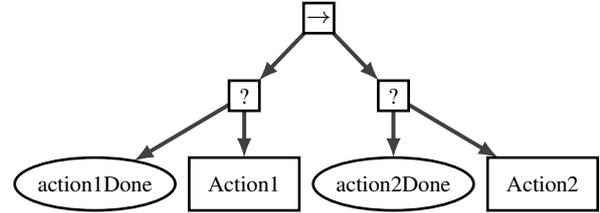
These always select the same actions given the same input, so this version is still not reactive. However the additional variables constitute implicit memory, so the non-reactiveness has essentially become less explicit. The operation of the variables $Action1Done$ and $Action2Done$ is no longer defined in any specific part of the structure, unlike when using a node with memory directly.

\subsubsection{Styles}

\emph{Styles} are another BT extension designed to operate between layers of the architecture, and can be used to introduce memory.

Styles~\cite{isla2015handling} are a BT extension designed to add more complex behavior to large BTs by modifying them in runtime. Specifically, the BT may be in one of several \emph{styles}, each of which disables some subtrees of the tree. Otherwise, while in a given style, the BT operates as usual but simply ignores the disabled subtrees. In different styles the BT may react slightly differently. This was suggested as use for coding the behavior of groups of BTs in~\cite{isla2015handling}, where the individual BTs are influenced by the current style of the group. The switching between styles may itself be done by a BT, or another architecture like a FSM. If the switching between styles is itself governed by a reactive architecture, then the BT layer will still always be reactive. However if the styles are controlled by a non-reactive architecture, then the overall behavior can be non-reactive.

Styles are an interesting example because, though the BT is modified at runtime and so may react differently to the same input (violating reactiveness), in any specific individual style none of the BT principles are violated. Styles are an example of an extension which operates on the layer \emph{above} the BT layer, so the use of styles is just one way of implementing a layer above BTs. The advantage of using a style on this layer is that much of the information stored in the BT is reused with styles, reflecting the fact that we often only need variations on behaviors not entirely distinct behaviors. Styles thus, are useful where they are applicable, for improving transparency of the relationships between layers.

\subsubsection{Decorators}  \label{sec:decorators}

So far, we have omitted mention of the Decorator node commonly used for BTs. This is largely because its extreme generality make it difficult to summarise, and must be analysed for particular Decorator instances. Here we very briefly discuss a few of these and apply some of the principles discussed above to their use.

\emph{Negation:} Negation is a Decorator node which returns Running if its child returns Running, Success if its child returns Failure and Failure if its child returns Success. This does not violate any principles of BT use. If its child is reactive then it is reactive and it is modular as it preserves the BT interface. What's more, it is extremely simple and interacts well with the other operators (preserving the symmetry of Success and Failure and providing an analogue of De Morgan's laws, as shown in~\cite{biggar2020framework}) making it very readable.

\emph{Run until Success (/Failure)} This Decorator ticks its child and returns the child's return value until the child returns Success, after which the Decorator returns Success always. It is easy to see that as its behavior is dependent on past return values, it violates reactivity. However, its behavior is at least fairly simple and constrained as the reactiveness is only lost after the child returns Success, minimising the impact of this loss of reactiveness. Again we show in~\cite{expressiveness} that BTs with such decorators are still strictly less expressive than FSMs.

\emph{Run $n$ times:} This Decorator ticks its child $n$ times, and after this point returns Success always. As before, it is not reactive, and can be more opaque than the previous, because the number of internal ticks does not necessarily correspond to any outwardly visible property.

\subsection{Modularity}

\subsubsection{Optimal gameplay} \label{sec:optimisation}

Behavior Trees were introduced in game AI to control the behavior of Non-Player Characters. In that context, one desires behavior which is realistic and complex, and which may be required to compete against the player where necessary. How much of this computation should be contained in the BT layer?

\begin{figure}
    \centering
    \begin{tikzpicture}
\clip (0,0) rectangle (7.0,3.0);
\Vertex[x=4.428,y=2.600,size=0.4,opacity=0,label=?,fontscale=1.2,shape=rectangle,style={}]{1}
\Vertex[x=3.460,y=1.571,size=0.4,opacity=0,label=$\rightarrow$,fontscale=1.2,shape=rectangle,style={}]{2}
\Vertex[x=1.562,y=0.400,size=0.7,opacity=0,label=Ghost Close,fontscale=1.2,shape=ellipse,style={minimum width = width("ghost close")+2ex}]{3}
\Vertex[x=5.357,y=0.400,size=0.7,opacity=0,label=Avoid Ghost,fontscale=1.2,shape=rectangle,style={minimum width = width("Avoidsghost")+2ex}]{4}
\Vertex[x=5.438,y=1.571,size=0.7,opacity=0,label=Eat Pills,fontscale=1.2,shape=rectangle,style={minimum width = width("Eatspills")+2ex}]{5}
\Edge[,fontscale=1.2,Direct](1)(2)
\Edge[,fontscale=1.2,Direct](1)(5)
\Edge[,fontscale=1.2,Direct](2)(3)
\Edge[,fontscale=1.2,Direct](2)(4)
\end{tikzpicture}
    \caption{A simple BT for playing Pac-Man.}
    \label{fig:pacman1}
\end{figure}
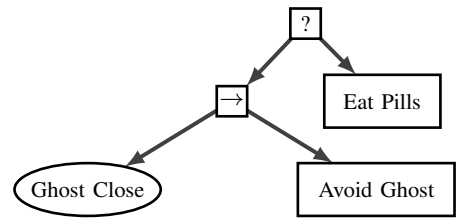

To begin, consider the BT in Figure~\ref{fig:pacman1} for playing the game Pac-Man.
This BT produces fairly trivial greedy behavior, but it is only a very simple tree. Can we do better with a more complex tree?

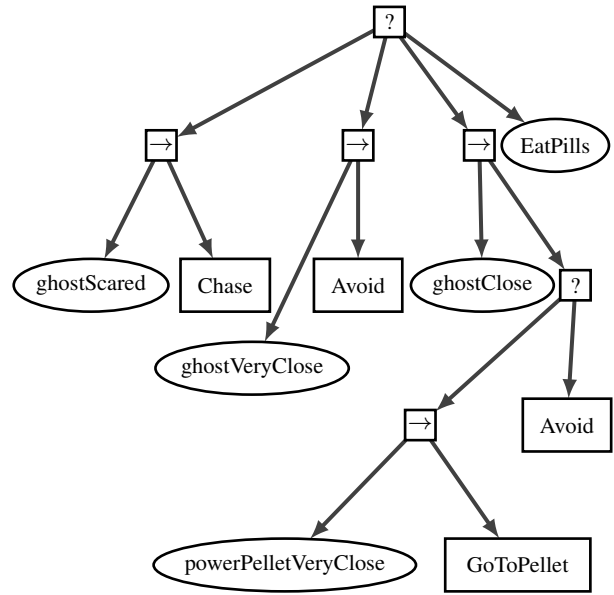
\begin{figure}
    \centering
    \begin{tikzpicture}
\clip (0,0) rectangle (8.0,8.0);
\Vertex[x=5.018,y=7.600,size=0.4,opacity=0,label=?,fontscale=1.2,shape=rectangle,style={}]{1}
\Vertex[x=2.018,y=5.964,size=0.4,opacity=0,label=$\rightarrow$,fontscale=1.2,shape=rectangle,style={}]{2}
\Vertex[x=1.094,y=4.109,size=0.7,opacity=0,label=ghostScared,fontscale=1.2,shape=ellipse,style={minimum width = width("ghostScared")+2ex}]{3}
\Vertex[x=2.858,y=4.109,size=0.7,opacity=0,label=Chase,fontscale=1.2,shape=rectangle,style={minimum width = width("Chase")+2ex}]{4}
\Vertex[x=4.606,y=5.964,size=0.4,opacity=0,label=$\rightarrow$,fontscale=1.2,shape=rectangle,style={}]{5}
\Vertex[x=3.212,y=3.009,size=0.7,opacity=0,label=ghostVeryClose,fontscale=1.2,shape=ellipse,style={minimum width = width("ghostVeryClose")+2ex}]{6}
\Vertex[x=4.606,y=4.109,size=0.7,opacity=0,label=Avoid,fontscale=1.2,shape=rectangle,style={minimum width = width("Avoid")+2ex}]{7}
\Vertex[x=6.2,y=5.964,size=0.4,opacity=0,label=$\rightarrow$,fontscale=1.2,shape=rectangle,style={}]{8}
\Vertex[x=6.24,y=4.109,size=0.7,opacity=0,label=ghostClose,fontscale=1.2,shape=ellipse,style={minimum width = width("ghostClose")+2ex}]{9}
\Vertex[x=7.464,y=4.109,size=0.4,opacity=0,label=?,fontscale=1.2,shape=rectangle,style={}]{10}
\Vertex[x=5.424,y=2.255,size=0.4,opacity=0,label=$\rightarrow$,fontscale=1.2,shape=rectangle,style={}]{11}
\Vertex[x=3.642,y=0.400,size=0.7,opacity=0,label=powerPelletVeryClose,fontscale=1.2,shape=ellipse,style={minimum width = width("powerPelletVeryClose")+2ex}]{12}
\Vertex[x=6.706,y=0.400,size=0.7,opacity=0,label=GoToPellet,fontscale=1.2,shape=rectangle,style={minimum width = width("GoToPellet")+2ex}]{13}
\Vertex[x=7.353,y=2.255,size=0.7,opacity=0,label=Avoid,fontscale=1.2,shape=rectangle,style={minimum width = width("Avoid")+2ex}]{14}
\Vertex[x=7.2,y=5.964,size=0.7,opacity=0,label=EatPills,fontscale=1.2,shape=ellipse,style={minimum width = width("EatPills")+2ex}]{15}
\Edge[,fontscale=1.2,Direct](1)(2)
\Edge[,fontscale=1.2,Direct](1)(5)
\Edge[,fontscale=1.2,Direct](1)(8)
\Edge[,fontscale=1.2,Direct](1)(15)
\Edge[,fontscale=1.2,Direct](2)(3)
\Edge[,fontscale=1.2,Direct](2)(4)
\Edge[,fontscale=1.2,Direct](5)(6)
\Edge[,fontscale=1.2,Direct](5)(7)
\Edge[,fontscale=1.2,Direct](8)(9)
\Edge[,fontscale=1.2,Direct](8)(10)
\Edge[,fontscale=1.2,Direct](10)(11)
\Edge[,fontscale=1.2,Direct](10)(14)
\Edge[,fontscale=1.2,Direct](11)(12)
\Edge[,fontscale=1.2,Direct](11)(13)
\end{tikzpicture}
    \caption{A more complex BT for playing Pac-Man.}
    \label{fig:pacman2}
\end{figure}

The simplest way to construct a more complex strategy is to check for more conditions, refining the partition of the state space induced by this reactive architecture. An example of such an improvement is the BT in Figure~\ref{fig:pacman2}.
This BT produces a slightly cleverer behavior, being able to decide whether to flee or to attempt to reach a Power Pellet on the basis of the distance to the nearest Ghost. However, it also does not produce optimal play, as it only considers the distance to ghosts, and not for instance whether Pac-Man will become trapped.

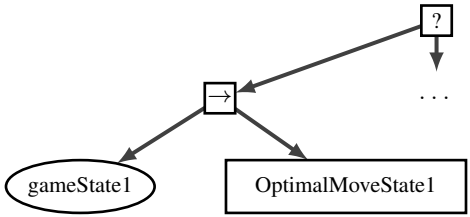
\begin{figure}
    \centering
    \begin{tikzpicture}
\clip (0,0) rectangle (7.0,3.0);
\Vertex[x=6.338,y=2.600,size=0.4,opacity=0,label=?,fontscale=1.2,shape=rectangle,style={}]{1}
\Vertex[x=3.490,y=1.571,size=0.4,opacity=0,label=$\rightarrow$,fontscale=1.2,shape=rectangle,style={}]{2}
\Vertex[x=1.653,y=0.400,size=0.7,opacity=0,label=gameState1,fontscale=1.2,shape=ellipse,style={minimum width = width("gameState1")+2ex}]{3}
\Vertex[x=5.147,y=0.400,size=0.7,opacity=0,label=OptimalMoveState1,fontscale=1.2,shape=rectangle,style={minimum width = width("OptimalMoveState1")+2ex}]{4}
\Vertex[x=6.338,y=1.571,Pseudo,size=0.7,opacity=0,label=$\dots$,fontscale=1.2,shape=ellipse,style={minimum width = width("anything")+2ex}]{5}
\Edge[,fontscale=1.2,Direct](1)(2)
\Edge[,fontscale=1.2,Direct](1)(5)
\Edge[,fontscale=1.2,Direct](2)(3)
\Edge[,fontscale=1.2,Direct](2)(4)
\end{tikzpicture}
    \caption{An optimal BT for a game such as Pac-Man.}
    \label{fig:optimalbt}
\end{figure}

One might be inclined to think, given the discussion above, that the limitations here are a result of the reactiveness of BTs, but in essence it is a limitation of their finite presentation. In fact, at least in theory, reactiveness is not limiting for this kind of AI. Pac-Man is discrete, and there are only a finite number of game states defined by the locations of all Ghosts, Pills and Pac-man. Further, the optimal move in Pac-Man only ever depends on the current world state---in other words, an optimal Pac-Man policy will be reactive.\footnote{In fact, a number of games have reactive optimal policies (despite possibly being very difficult to play optimally). As classic AI examples, both chess and Go have reactive optimal policies, if one ignores the few rules in either game which do not depend solely on the current board state. These include the \emph{en passant}, castling and threefold-repetition rules in chess, and the \emph{ko} rule in Go.} Suppose such an optimal policy was given to us. By constructing a tree with all possible game states, and associating each with its optimal $Up,Down,Left,Right$ move from this policy, we can construct a BT in the form of Figure~\ref{fig:optimalbt}.
 Such a BT is both completely reactive and optimal, but also completely unrealistic. Firstly, the assumption of having an optimal policy given is very strong or, for most games, impossible. Secondly, constructing such a tree is well beyond the realm of feasibility for even games like Pac-Man.
 Thirdly, and most importantly for this paper, this approach violates the \emph{modularity} of BTs. As the specificity of the actions and conditions grow, their reusability reduces. The subtrees of the tree begin to correspond less and less to identifiable behaviors. It is understandable that there would be a trade-off between optimal behavior and transparency. Optimisation, being computationally intensive, makes it infeasible for an outside observer to predicts the agent's behavior any faster than that agent can calculate its own behavior. Striking a balance between these competing goals in robotics can be difficult. In general though, we suggest that where optimisation-based policies are required they should be separated into individual actions and implemented at a layer below the BT. In this way, the BT layer is still readable and modular, and adds reactiveness on top of these policies.

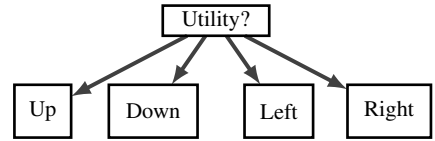
\begin{figure}
    \centering
    \begin{tikzpicture}
\clip (0,0) rectangle (7.0,2.0);
\Vertex[x=3.500,y=1.600,size=0.4,opacity=0,label=Utility?,fontscale=1.2,shape=rectangle,style={minimum width=1.4cm}]{1}
\Vertex[x=1.222,y=0.400,size=0.7,opacity=0,label=Up,fontscale=1.2,shape=rectangle,style={minimum width = width("Up")+2ex}]{2}
\Vertex[x=2.674,y=0.400,size=0.7,opacity=0,label=Down,fontscale=1.2,shape=rectangle,style={minimum width = width("Down")+2ex}]{3}
\Vertex[x=4.326,y=0.400,size=0.7,opacity=0,label=Left,fontscale=1.2,shape=rectangle,style={minimum width = width("Left")+2ex}]{4}
\Vertex[x=5.778,y=0.400,size=0.7,opacity=0,label=Right,fontscale=1.2,shape=rectangle,style={minimum width = width("Right")+2ex}]{5}
\Edge[,fontscale=1.2,Direct](1)(2)
\Edge[,fontscale=1.2,Direct](1)(3)
\Edge[,fontscale=1.2,Direct](1)(4)
\Edge[,fontscale=1.2,Direct](1)(5)
\end{tikzpicture}
    \caption{A translation of Figure~\ref{fig:optimalbt} using a Utility node.}
    \label{fig:util}
\end{figure}

Another proposed solution to such cases is a BT extension called the \emph{Utility node}. Utility nodes operate like a Fallback node, except that an internal optimisation rearranges the order of the children at runtime, in order of highest to lowest `utility' in that circumstance. With this tool, we could in principle restructure the above optimal Pac-Man BT as the BT in Figure~\ref{fig:util}.
This BT is no more useful than the previous, but this is not surprising because there is no straightforward way to represent this optimal policy. However, in more practical cases Utility BTs can provide useful insight into architectures performing optimisation. If an observer can inspect the `utility scores' being generated at runtime, then this approach provides some readable insight into the underlying optimisation, while not necessarily being reactive. If subtrees and actions can calculate their own utility scores internally (thus maintaining their conceptual cohesiveness), this approach can still be modular.

This Pac-Man example was adapted from~\cite{btbook}.

\subsubsection{BTs on atomic actions} \label{sec:atomic actions}

Thus far, our discussion has been largely agnostic to how many layers of control exist above or below the BT layer. However the previous section noted that to construct BTs that display closer to `optimal' behavior, the number of conditions required grows exponentially. The section concluded by arguing that an optimisation computation was better placed at a lower layer than the BT. In this section we dig further into this problem.

In principle, BTs expressiveness does not depend on the content of the actions that are given to it. These could be extremely complex behaviors, or extremely low-level signals. In practice though, it is often the case that the limitations of BT expressiveness, as discussed in~\cite{expressiveness} are far more apparent for low-level actions. A common example of such use is BTs for discrete gameplay~\cite{colledanchise2018learning,nicolau2016evolutionary,perez2011evolving} or BTs evolved from a finite set of atomic actions and conditions~\cite{christensen2016evolving,lim2010evolving,perez2011evolving}. For example, consider a BT operating in a discrete grid world. Some obvious simple actions in this world might be Step Forward, Turn Left, Turn Right, Pick Up Object, etc. Alternatively, more `complex' actions such as Explore, Go to Region X, Avoid Enemy, etc could be used. How should we decide which of these sets of actions is most appropriate for the BT to operate upon? Essentially, how many layers of the architecture should lie between the inputs from this discrete world and the BT layer?

To answer this question, we must first observe that in general the layers of the control architecture provide layers of abstraction, and corresponding reductions in the size of the information passed upwards. At the lowest layer, input data is largest in volume, and for real-world robots often close to continuous. On the other hand, the data passed to the highest level is a highly abstract description of the input and lower layers. It does not make sense for a layer to receive input data, process it, then output more data than it received.

This means that, even for a discrete world such as the example above, there is a state-explosion problem for the lowest layers. Suppose for the sake of argument that the input is given by the states of the world grid in the $5\times 5$ square surrounding the agent, and that each square can be in one of 5 states. This gives $5^{25}$ possible inputs. A BT attempting to reason in such a world then will suffer from one of two problems. The first option is that it responds differently to large number of input states. This generally causes it to grow to exponential size, which is not modular, not readable, and largely impractical. It is also likely to violate our assumption that action selection took negligible time on the world time-scale, as selecting from this number of inputs could be computationally intensive. In addition, recall that when justifying why reactiveness was desirable, we pointed out that it made behavior intuitive and easy to understand. If an architecture is reactive but has $5^{25}$ possible inputs, it is essentially impossible for an observer to discern this reactiveness. 

The other alternative for a low-layer BT is that it significantly abstracts the input space, by constructing a small number of conditions on this input. This allows BTs of reasonable size and reactiveness to be built. However, this abstraction throws away a significant amount of information, which can no longer be used by the BT or any layer built above it. This greatly limits the reasoning that can be done. As a grid-world example, suppose we construct conditions Wall Left/Right, indicating there is a wall somewhere in one of the closest two squares to the left/right of the agent, but no other conditions exist for sensing Walls. If the condition $\text{Wall Left}\land\text{Wall Right}$ is true, we have lost the ability to compute on which side the wall is closer to the agent.

What's more, the restrictions on expressiveness that are induced by the criteria of reactiveness (discussed fully in~\cite{expressiveness}) become more stringent at this level of atomicity. In fact, some simple higher-level behaviors are impossible to construct using only a reactive architecture (such as a BT) from atomic actions such as Turn Left and Step Forward. Consider the problem of wall-following in a discrete world. We want a behavior which, if placed next to a wall, can follow that wall until it returns to its starting point. While such a behavior can be constructed from an FSM, we cannot with a BT. Consider the following sequence of steps, where we assume the agent has a $5\times 5$ field of vision with no other walls within this view.
\begin{figure*}
    \centering
    \begin{tikzpicture}
\clip (0,0) rectangle (14.4,4);
\draw[step=0.8cm,opacity=0.2,very thin] (0,0) grid (14.4,4);

\draw (0.8,0) -- (0.8,1.6) -- (1.6,1.6) -- (1.6,0);
\draw (0.8,4) -- (0.8,2.4) -- (1.6,2.4) -- (1.6,4);
\node[circle,line width= 0.05cm,draw=black,minimum size=0.5cm] at (2,1.2) {$\uparrow$}; 
\draw[dashed] (2.4,0) -- (2.4,4);
\node[fill=white] at (1.2,3.5) {Step 0};

\draw (3.2,0) -- (3.2,1.6) -- (4,1.6) -- (4,0);
\draw (3.2,4) -- (3.2,2.4) -- (4,2.4) -- (4,4);
\node[circle,line width= 0.05cm,draw=black,minimum size=0.5cm] at (4.4,2) {$\uparrow$}; 
\draw[dashed] (4.8,0) -- (4.8,4);
\node[fill=white] at (3.6,3.5) {Step 1};

\draw (5.6,0) -- (5.6,1.6) -- (6.4,1.6) -- (6.4,0);
\draw (5.6,4) -- (5.6,2.4) -- (6.4,2.4) -- (6.4,4);
\node[circle,line width= 0.05cm,draw=black,minimum size=0.5cm] at (6.8,2) {$\leftarrow$}; 
\draw[dashed] (7.2,0) -- (7.2,4);
\node[fill=white] at (6,3.5) {Step 2};

\draw (8,0) -- (8,1.6) -- (8.8,1.6) -- (8.8,0);
\draw (8,4) -- (8,2.4) -- (8.8,2.4) -- (8.8,4);
\node[circle,line width= 0.05cm,draw=black,minimum size=0.5cm] at (8.4,2) {$\leftarrow$}; 
\draw[dashed] (9.6,0) -- (9.6,4);
\node[fill=white] at (8.4,3.5) {Step 3};

\draw (10.4,0) -- (10.4,1.6) -- (11.2,1.6) -- (11.2,0);
\draw (10.4,4) -- (10.4,2.4) -- (11.2,2.4) -- (11.2,4);
\node[circle,line width= 0.05cm,draw=black,minimum size=0.5cm] at (10,2) {$\leftarrow$}; 
\draw[dashed] (12,0) -- (12,4);
\node[fill=white] at (10.8,3.5) {Step 4};

\draw (12.8,0) -- (12.8,1.6) -- (13.6,1.6) -- (13.6,0);
\draw (12.8,4) -- (12.8,2.4) -- (13.6,2.4) -- (13.6,4);
\node[circle,line width= 0.05cm,draw=black,minimum size=0.5cm] at (12.4,2) {$\downarrow$}; 
\node[fill=white] at (13.2,3.5) {Step 5};

\end{tikzpicture}
    \caption{A reactive agent becoming stuck in a loop while attempting to follow a wall. Step 5 is identical to Step 1 from the agent's perspective, so the agent turns back through the wall. The arrow denotes the direction the agent is facing.}
    \label{fig:wall_following}
\end{figure*}
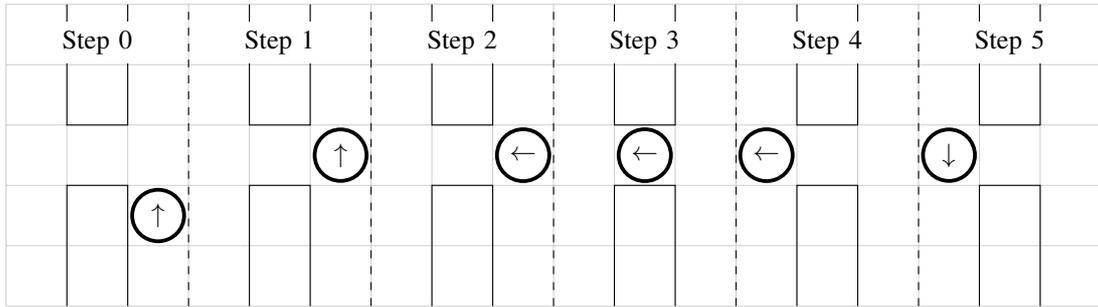
In this example, the agent is following the lower wall, when it reaches a passage. It passes through the passage (Steps 1-5) but upon reaching Step 5 the input is identical to the state of Step 1, from the agent's perspective. Thus, the agent returns and passes back through the passage, and so becomes caught in a loop. While this is not a proof, it is possible to construct this or similar arguments formally. We omit such an argument because the limitations on expressiveness of reactive architectures is covered in~\cite{expressiveness}.

While FSMs can be used to construct a wall-following algorithm, many of the arguments in this section also apply to FSMs, and indeed FSMs are rarely used at the lowest levels of abstraction in practice. We would not use an FSM to perform robotic vision, to extrapolate from sensor data or to calculate a shortest path. We should therefore not aim to construct BTs for these purposes.

\subsubsection{Encapsulation} \label{sec:encapsulation}

The points thus far have argued that we should use BTs where we can preserve their modularity and reactiveness. However, we know~\cite{nilsson1993teleo,btbook,modularity} that these properties are also critical for Teleo-reactive programs and Decision Trees. How then should we decide between structuring a reactive layer as a BT or another of these reactive architectures? TRs and DTs are arguably conceptually simpler than BTs, so a priori they seem more appealing. For instance, we can emulate a BT with a Decision Tree by representing action return conditions as explicit predicates in the DT, which are then linked up with appropriate branching. This DT makes exactly the same action selections as the original BT, so what is the difference between these approaches?

We postulate that the BT representation has the advantage of \emph{encapsulating its own metadata}, in the form of return values. This aspect of BT's modularity can be considered to take inspiration from the object-oriented programming paradigm. DTs use predicates on the current input to select the most `appropriate' action. These predicates are often naturally related with specific actions, in every DT in which that action is used. Reusing an example from Section~\ref{sec:modellingreturnvalues}, consider the action Grasp Object. In choosing when to select this in a DT, the information of whether the robot is already grasping an object, or if there is no object to grasp, will almost always be relevant. It is for this reason that we associated these two conditions permanently to this particular action, by assigning them as the Success and Failure conditions. Given this information is naturally related to this action, it should be stored within that action so that the action can be easily reused. In a DT, if we explicitly express these Success and Failure conditions as predicates in the tree, they have been become disconnected from their respective action. Because this information is not self-contained, when using the action Grasp Object in another DT, we must either store this relevant information somewhere or have it be lost. For BTs, this information is stored in the tree and accessed by the operators, allowing complex trees with complex return conditions to be derived and stored for later reuse. Recalling Section~\ref{sec:modellingreturnvalues} on modelling return conditions, we conclude that to maximise reuse we should store as return values all the information that relates closely to that action.

\subsubsection{Reuse} \label{sec:reuse}

The findings of the above section suggest that we should associate all clearly related values with actions in order to maximise their ability to be reused---noting that doing so also maximises the ability of trees derived from these actions to be reused, in accordance with the principle of modularity.

However, consider a Teleo-reactive program, which itself is a modular~\cite{modularity} control architecture. TRs, like BTs, encapsulate data associated with actions to some degree, by assigning each action a precondition. When comparing BTs to TRs, as done in~\cite{colledanchise2016behavior,btbook}, this condition is usually considered as either a Success or Failure condition. The important thing to note is that in general it is perfectly reasonable to interpret it in either sense. In addition, recall that in Section~\ref{sec:modellingreturnvalues} we argued that not all actions have a  natural sense of Succeeding or Failing. The resultant point is that we should not assign any particular meaning to Success and Failure. Instead, like the precondition of an action in a TR, they are better thought of as two useful pieces of metadata which can be accessed for the purposes of action selection. The BT framework is set up in general to emphasise symmetry between these two values. This symmetry is useful and makes reasoning about BTs intuitive, which we argue means that we should not assign any particular meaning or criterion to one value but not the other.

By doing this, we are allowing for increased reuse of actions and behaviors across multiple architectures. For instance, actions with preconditions used in TRs can be used without modifications in BTs, simply by interpreting the precondition in a fixed way as Success or Failure. Likewise, an action which never returns either Success or Failure can be used in any of BTs, TRs or DTs. This means in addition that TRs and DTs can be used as actions within BTs, for instance by using the translation of~\cite{generalise} (nesting architectures is discussed formally in~\cite{modularity}).

We can further improve modularity by removing all reliance on the `Running' return value. Instead, we assume as in~\cite{modularity}, that there exists a return value function attached to each action, which gives a value in any input. For an action in a BT, two of these values may be interpreted as Success and Failure, but there could be any number of other values. We shall now interpret `Running' for this action as \emph{any return value not handled by the structure} \footnote{In some early forms of BTs there was no Running return value~\cite{mateas2002behavior}, and instead actions executed one by one until they returned Success or Failure. That is not reactive, and very different from what we mean here. We assume there is at least one other value that can be returned, and that a value is returned at every time step. We are simply allowing information to be passed that would otherwise be hidden by a `Running' value.}. A priori, nothing has changed with regard to the operation of this action in a BT, because the tree responds appropriately to the values Success and Failure and any other values are interpreted as Running. However, this definition allows any action to be used in any of these architectures---we no longer need to make assumptions about the number of provided return values. For instance, in a decision tree any value is interpreted as Running, as the tree does not handle any return values. Similarly, we could use any action in a TR, by choosing a value to represent its precondition and ignoring all other values returned. This now allows BTs to be used as subtrees in DTs, or as individual actions within a TR, maximising the ability to reuse complex behavior. By assigning return values in this way, we allow actions to be stored with their metadata independent of which architecture(s) they will eventually become a part of.

This requires something of a mental shift from the usual BT interpretation, because the somewhat loaded terms Success and Failure suggest that that action cannot be selected as it has already `Succeeded' or `Failed'. However, as we already saw in Section~\ref{sec:modellingreturnvalues}, we shouldn't think of these values in the past tense as following the `completion' of an action. In a reactive architecture then, these values may not correspond to intuition. As in the example action Grasp Object, it is not \emph{unreasonable} to attempt to grasp an object while already holding one, it is merely \emph{trivial}; the resultant action has no effect. The condition that we are already holding an object, which we assigned as the metadata associated with the value `Success', is used to determine that in this situation it may be desirable to select a different action. In formally defining BTs in Section~\ref{sec:asms}, we already allowed for actions to be selected even if they returned a non-Running value because this allowed architectures to be composed. Here we have extended this idea.

To cap off this discussion, we should recall the result of the previous section, which is that we should assign to an action precisely those return values that are naturally associated with it in all contexts. These may or may not directly correspond to Success and Failure. The results here have shown that we can still apply that action to all of these architectures. Further, all of those architectures can be used as actions within each other without modification. If for instance we construct a TR, but then later decide to embed this as a subtree in a BT, this can easily be done, and vice versa.

\section{Generalising Behavior Trees: the $k$-BTs}
\label{sec:kbts}

The previous two subsections allowed us to conclude that we should assign all appropriate return values to actions, to maximise their modularity. However, BTs can only ever access two of these values, and TRs and DTs can only access one or zero respectively. In this section we show how a simple generalisation of BTs extends their benefits to a broader class of behaviors which can react to any number of return values. We call these the \emph{generalised Behavior Trees} or \emph{$k$-BTs} and we show, as in the previous section, that BTs can be used without modification as subtrees of $k$-BTs. Further, the normal BTs are exactly the 2-BTs and the TRs are the 1-BTs. Finally, we show how $k$-BTs provide natural solutions to some of the examples given in Section~\ref{sec:practical}.

To introduce these objects, first recall that BTs are trees with two control flow nodes, corresponding to the specific values Success and Failure (note that we are omitting the Decorator and Parallel nodes, for which analogous constructions could be made if necessary). The $k$-BTs will be defined as trees with $k$ control flow nodes. We will choose $k$ return values, which we shall write as the values $\{1,\dots,k\}$, and we shall associate a control flow node with each. Consistent with the previous section, there is no explicit `Running' value.

Formally, we define the $k$-BTs as follows. We will denote the $k$ control flow nodes by $*_1,\dots,*_k$. Fix an ordered rooted tree with $\ell$ leaves labelled by actions and internal nodes labelled by these control flow nodes. When a control flow node $*_i$ is ticked, it ticks each of its children from left to right, until some child returns a value other than $i$, at which point that control flow node returns that same value. More formally, each operator $*_i$ is defined by the following rules. If $*_i$ has one child $\alpha$, select that child. Otherwise, for an input $x\in\Inp$, if $*_i$ has children $c_1,\dots,c_n$,
\begin{itemize}
    \item Begin at $c_1$. Select $c_j$ if $c_j$ does not return $i$ in $x$.
    \item Otherwise, repeat for $c_{j+1}$.
    \item Select $c_n$ if no previous are selected
\end{itemize}
The formal definition of the overall $k$-BT is built by recursively combining these, in exactly the same way that BTs are built from the definitions of Sequence and Fallback.

Note now that if only two control flow nodes are used (which we assume for simplicity correspond to values 1 and 2) we can interpret them as Sequence and Fallback and determine that the 2-BTs are exactly the normal BTs. Similarly, if the value 1 (or any fixed value) corresponds to the precondition of an action being false, then the 1-BTs are exactly the TRs. Hence we have constructed a family of modular reactive control architectures which include BTs and TRs. In fact, it is easy to see that the $k$-BTs inherit the reactiveness, modularity and readability of BTs as they inherit BT's intuitive presentation as a tree. Finally, note that the `$k$' in $k$-BTs is a placeholder, which we can be replaced with the 2-BTs or 3-BTs as above, but when left unspecified we assume $k$ is any fixed finite value.

\subsection{Use cases for $k$-BTs} \label{sec:kbtusage}
Since the 2-BTs are the standard BTs, $k$-BTs are clearly applicable for any situation where BTs are used. The previous sections have also established that the existence of $k$-BTs is at least a consistent consequence of modularity, reactiveness and readability. Here we show that the $k$-BTs are practically useful, in that they provide a natural way of extending BTs in certain cases where they are otherwise unwieldy. As examples, we will we show here how $k$-BTs provide a possible solution to the challenges of Sections~\ref{sec:modellingreturnvalues} (Modelling Success and Failure),~\ref{sec:conditions requiring action} (Knowledge Versus Memory).

 The way that Sequence and Fallback handle the values Success and Failure from their children can be thought of in terms of exceptions in programming. These values then are two distinct exceptions which can be thrown by a program, with Sequence and Fallback acting as code blocks which catch and process the respective exceptions. Adding further control flow nodes is no different than catching more exceptions. The `thrown error' analogy provides insight to an obvious use case for $k$-BTs: failure handling. In~\cite{btbook}, the authors argue that failure handling is easier in BTs than TRs, and significantly easier than in DTs, as more return values can be handled. the $k$-BTs make this yet more streamlined, essentially by allowing for numerous distinct error codes which can be handled differently. Failure conditions can now be broken into any number of natural failure cases and assigned distinct return values. Of course, as we do not consider either Success or Failure to have inherent meaning, the additional return values can equally be interpreted as different types of Success.

These values can also be though of as degrees of Success (or Failure). For instance, consider a BT controlling an agent with an action which succeeds when some number of items are in a goal region. In such a case, it could be useful to have metadata storing a `partially successful' value if some number are within the goal region. This same idea could be extended to risk, with various return values indicating the probability of successful of unsuccessful completion. This type of reasoning is useful for tasks where overall Success/Failure is unusual or not critical, and where a `reasonable' performance is acceptable.

We need not solely interpret the additional values as variants of Success and Failure. For example, consider the discussion in Section~\ref{sec:conditions requiring action}, where a robot action Unlock Door returns Success if the door is unlocked and Failure if there is no key. However, its action will depend on whether the knowledge of whether the door is locked is stored within its representation of the world. In this case, we could add an additional value `Unknown', which is returned when the robot is uncertain about the status of the door. This can be either handled in the tree, allowing us to not execute this action if the robot is uncertain, or ignored in the tree, in which case it is treated as `Running' and the action is still selected. In both cases, `Unknown' is not merely a variation on Success or Failure but a concept which cannot be be obviously defined as either.

\section{Conclusions and Future Work} \label{sec:conclusions}

In this paper we have analysed how BTs can be used in a manner consistent with modularity and reactiveness. As far as achieving the goals of reusable and readable AI goes, there are significant directions for extension of this work. Firstly, the discussion of BT use cases was by no means comprehensive, and there are a number of additional cases worthy of discussion with regards to these principles. Secondly, many of the ideas of this work applied beyond BTs, and similar work exploring other architectures seems worthwhile.

\bibliographystyle{IEEEtran}
\bibliography{references.bib}
\end{document}